\documentclass[11pt,letterpaper]{article}
\usepackage[letterpaper]{geometry}
\usepackage{acl2012}
\usepackage{times}
\usepackage{latexsym}
\usepackage{amsmath}
\usepackage{amssymb}
\usepackage{multirow}
\usepackage{url}
\makeatletter
\newcommand{\@BIBLABEL}{\@emptybiblabel}
\newcommand{\@emptybiblabel}[1]{}
\makeatother
\usepackage[hidelinks]{hyperref}
\usepackage{graphicx}
\usepackage{adjustbox}

\title{Non-Projective Dependency Parsing via Latent Heads Representation (LHR)}

\author{Matteo Grella \\
  Turin, Italy \\
  {\tt matteogrella@gmail.com} \\\And
  Simone Cangialosi \\
  Turin, Italy \\
  {\tt sm.cangialosi@gmail.com} \\}

\date{}

\begin{document}
\maketitle
\begin{abstract}
In this paper we introduce a novel approach based on a bidirectional recurrent autoencoder to perform globally optimized non-projective dependency parsing via semi-supervised learning. The syntactic analysis is completed at the end of the neural process that generates a Latent Heads Representation (LHR), without any algorithmic constraint and with a linear complexity. The resulting ``latent syntactic structure'' can be used directly in other semantic tasks. The LHR is transformed into the usual dependency tree computing a simple vectors similarity. We believe that our model has the potential to compete with much more complex state-of-the-art parsing architectures.
\end{abstract}

\begin{figure*}[h]
\begin{center}
\includegraphics[scale=2.00]{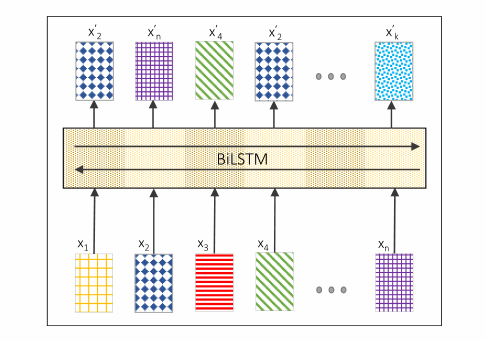}
\end{center}
\vspace*{-1em}
\caption{\small The bidirectional recurrent autoencoder used by our model is trained to associate to each input vector \texttt{x\textsubscript{i}} another input vector \texttt{x\textsubscript{j}} (i $\neq$ j). The output vectors of the BiLSTM pass through a Feedforward network that reduces their size to the one of the input vectors. In the example, for both \texttt{x\textsubscript{1}} and \texttt{x\textsubscript{4}} the autoencoder is trained to reconstruct \texttt{x\textsubscript{2}}. From the point of view of the dependency parsing, this means that \texttt{x\textsubscript{1}} and \texttt{x\textsubscript{4}} share the same head. In the picture corresponding vectors are represented with the same texture and color. }
\label{fig:schema_autoencoder}
\end{figure*}

\section{Introduction}

Dependency parsing is considered to be a fundamental step for linguistic processing because of its key importance in mediating between linguistic expression and meaning. The task is rather complex as the natural language is implicit, contextual, ambiguous and often imprecise. Recent data-driven deep-learning techniques have given very successful results in almost all the natural language processing tasks, including dependency parsing, thanks to their intrinsic ability to handle noisy inputs and the increased availability of training resources; see \newcite{goldberg-methods} for an introduction. \par

Generally speaking, modern approaches to dependency parsing can be categorized into graph-based and transition-based parsers \cite{kubler2008dependency}. Most neural dependency parsers take advantage of neural networks only for features extraction, using those features to boost traditional parsing algorithms and to reduce the need for feature engineering.\footnote{The ``old school'' of state-of-the-art parsers relied on hand-crafted feature functions \cite{zhang11acl}.} Starting from \newcite{chen2014fast} there has been an increase of sophisticated neural architectures for features representation (\newcite{dyer2015transitionbased}, \newcite{ballesteros2016dynamic}, \newcite{kiperwasser2016ef}, \newcite{dozat2016deep}, \newcite{liu2017encoder}, just to name a few). Many of them use recurrent neural networks. In particular, \newcite{kiperwasser2016simple} were the first who demonstrated the effectiveness of using a conceptually simple BiLSTM \cite{graves2008supervised,irsoy2014opinion} to reduce the features to a minimum \cite{shi2017fast}, achieving state-of-the-art results in both transition-based and graph-based approaches.\footnote{A BiLSTM (bidirectional LSTM) is composed of two LSTMs, $\textsc{LSTM}_{F}$ and $\textsc{LSTM}_{R}$, one reading the sequence in its regular order, and the other reading it in reverse.} \par

However, despite deep neural networks have proven to be successful in capturing the relevant information for the syntactic analysis, their use is still auxiliary to the traditional parsing algorithms. For instance, transition-based parsers use the neural components to predict the transitions, not directly the syntactic dependencies (arcs). Graph-based parsers, by contrast, use them to assign a weight to each possible arc and then construct the maximum spanning tree \cite{mst}. \par 

In this paper we introduce a novel approach based on a bidirectional recurrent autoencoder to perform globally optimized dependency parsing via semi-supervised learning. The syntactic analysis is completed at the end of the neural process that generates what we call \textbf{Latent Heads Representation} (\textbf{LHR}), without any algorithmic constraint and with a linear complexity. The resulting ``latent syntactic structure'' can be used directly for other high-level tasks that benefit from syntactic information (i.e. sentiment analysis, sentence similarity, neural machine translation). \par We use a simple decoder to transform the \textit{LHR} to the usual tree representation computing a vector similarity, with a quadratic complexity. \par 

An interesting property of our model compared to other approaches is that it handles unrestricted non-projective dependencies naturally, without increasing the complexity and without requiring any adaptation or post-processing.\footnote{This is particularly remarkable as discontinuities occur in most if not all natural languages: non-projective structures are attested also in more fixed word order languages, like English or Chinese.} \par 

The resulting parser has a very simple architecture and provides appreciable results without using any resource outside the tree-banks (with a baseline of \textbf{92.8\% UAS} on the English Penn-treebank \cite{penntb} annotated with Stanford Dependencies and non-gold tags). We believe that with some tuning our model has the potential to compete with much more complex state-of-the-art parsing architectures.

\begin{figure*}[h]
\begin{center}
\includegraphics[scale=0.50]{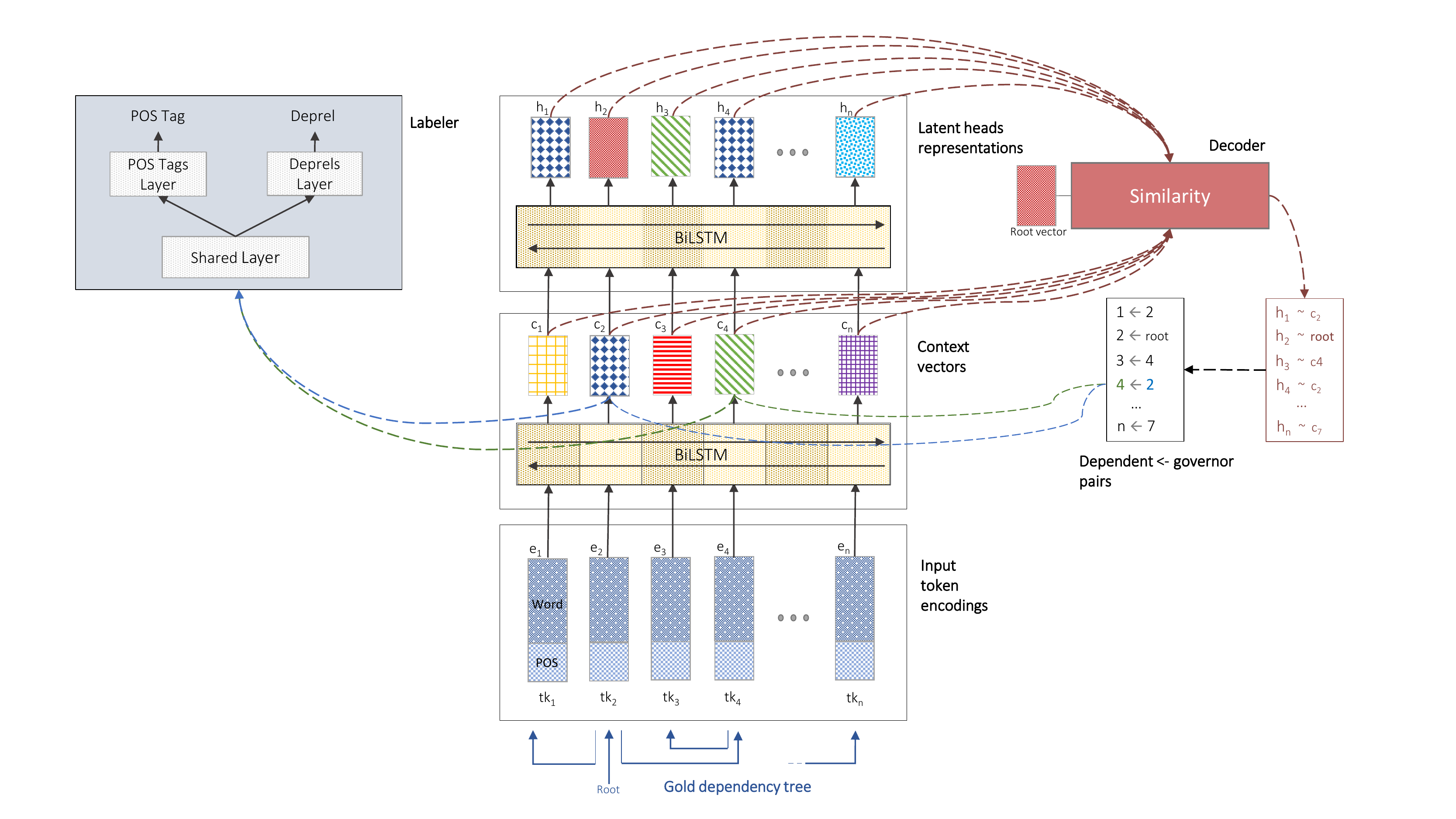}
\end{center}
\vspace*{-1em}
\caption{\small Illustration of the behavior of the neural model when parsing a sentence. The tokens of a sentence (\texttt{tk\textsubscript{1}...tk\textsubscript{n}}) are first transformed into a distributed representation (\texttt{e\textsubscript{1}...e\textsubscript{n}}) and then encoded into the \textit{context vectors} (\texttt{c\textsubscript{1}...c\textsubscript{n}}) by the \textit{context-encoder}. The \textit{context vectors} pass in turn through the \textit{heads-encoder} that predicts the \textit{latent-heads} (\texttt{h\textsubscript{1}...h\textsubscript{n}}).
The decoder finds the \textit{top token} \texttt{i-th} searching for the head \texttt{h\textsubscript{i}} most similar to the \textit{root vector}. Subsequently, for each token \texttt{i-th} its head token \texttt{j-th} is found such that \texttt{i $\neq$ j} and \texttt{c\textsubscript{j}} is the most similar to \texttt{h\textsubscript{i}}. After having found all the heads and removed all the cycles of the resulting dependency tree, the multi-tasking network assigns a deprel label and a POS tag to each token \texttt{i-th}, taking in input the concatenation of its \text{context vector} (\texttt{c\textsubscript{i}}) and that of its head (\texttt{c\textsubscript{j}}).}
\label{fig:schema_lhr}
\end{figure*}

\section{Our Approach}
\label{sec:approach}

\subsection{The Idea}
\label{sec:idea}

The dependency parsing consists of creating a syntactic structure of a sentence through word-to-word dependencies (\newcite{Tesniere:1959}, \newcite{Sgall:1986}, \newcite{Mel'cuk:1988}, \newcite{hudson90}) where words are linked by binary asymmetric relations called dependencies. \par

Like \newcite{zhang2016dependency} we formalize the dependency parsing as the task of finding for each word in a sentence its most probable head without tree structure constraints (see a comparison between ours and Zhang's model in Section \ref{sec:related}). \par

We propose a novel approach for dependency parsing based on the ability of the autoencoders \cite{rumelhart1985learning} to learn a representation with the purpose of reconstructing its own input.\footnote{While conceptually simple, autoencoders play an important role in machine learning and are one of the fundamental paradigms for unsupervised learning.} \par 

The gist of our idea is to use a bidirectional recurrent autoencoder (Figure \ref{fig:schema_autoencoder}) to reconstruct for each \textit{i-th} input another \textit{j-th} input of the same sequence, remaining in the same domain. In this way, we are able to train the network to create an approximate representation of the head of a given token.\footnote{Although in a different context, the work of \newcite{rama2016lstm} is the most similar to ours that we found. They use a LSTM sequence-to-sequence autoencoder to learn a deep words representation that can be used for meaningful comparison across dialects.} \par

As detailed in Section \ref{sec:unlabeled}, the input values of the autoencoder are not fixed as they are tuned on the basis of the errors of the autoencoder itself, involving a sort of information rebalancing able to converge. \par

Considering that the network learns its own suitable representation for both the input and the output by itself and that we give a ``teaching signal'' only (that is the index of the target vector of the head to reconstruct without imposing any particular representation) we consider our approach a semi-supervised one.

\subsection{Unlabeled Parsing}
\label{sec:unlabeled}

Our model (Figure \ref{fig:schema_lhr}) is composed of two BiRNN\footnote{In this paper, we use the term BiRNN to abstract the general concept of bidirectional recurrent network, not to refer to the specific model of Schuster and Paliwal (1997) who extended the simple recurrent network (Elman, 1990).} encoders which perform two distinct tasks. The first BiRNN (\textbf{\textit{context-encoder}}) receives in input the tokens of a sentence already encoded in a dense representation.\footnote{For example, the tokens encodings can be obtained by the concatenation of word and pos embeddings.} The \textit{context-encoder} encodes the input tokens into ``\textbf{\textit{context vectors}}'' that represent them with their surrounding context. Its contribute is crucial for the positional information that it adds to the input vectors, especially when a word occurs more than once in a sentence. \par

The \textit{context vectors} are in turn given as input to the second BiRNN (\textbf{\textit{heads-encoder}}) that acts as an autoencoder, transforming them into another representation that we call ``\textbf{\textit{latent heads}}''. The aim of the \textit{heads-encoder} is to associate each \textit{context vector} to the one that represents its head\footnote{The  output  vectors  of  the  BiLSTM  pass through a Feedforward network that reduces their size to the one of the input vectors.}. It is trained to minimize the difference between a \textit{context vector} and its representation as \textit{latent head}.\footnote{We use the mean squared error during the training phase and the cosine similarity during decoding.} During the training, the dependencies between dependent and governor tokens are taken from the gold dependency trees. \par

The mean absolute errors are propagated from the \textit{heads-encoder} all the way back, through the \textit{context-encoder} until the initial \textit{tokens embeddings} (which are trained together with the model). An optimizer is used to update the parameters according to the gradients. \par

The model is trained to predict the \textit{latent heads} of each sentence without a sequential order, generating all the tokens dependencies at the same time. So, we consider it globally optimized. \par

Thanks to the ability of the LSTMs \cite{hochreiter1997long} (or equivalent gated recurrent networks) to remember information for long periods, this method allows to recognize word-to-word dependencies among arbitrary positions in a sequence of words directly, without the need of any additional transition-based or graph-based framework. \par

To construct the dependency tree we use a decoder that finds the head (i.e. the governor) of each token searching for the \textit{context vector} most similar to its \textit{latent head} (excluding itself). The top token of the sentence is found before assigning the other heads, looking for the \textit{latent head} most similar to a reference vector used to represent the \textit{virtual root} (see Section \ref{sec:theroot}). \par 

At test time, we ensure that the dependency tree given in output is well-formed by iteratively identifying and fixing cycles with simple heuristics, without any loss in accuracy.\footnote{For each cycle, the fix is done by removing the arc with the lowest score and assigning to its dependent the node that maximizes its latent head similarity without introducing new cycles.} \par

Like \newcite{zhang2016dependency}, we empirically observed that during the decoding most outputs are already trees, without the need to fix cycles. It seems to confirm that in both our models the linear sequence of tokens itself is sufficient to recover the underlying dependency structure.

\subsection{Labeled Parsing}
\label{sec:labeled}

Up to now, we described unlabeled parsing. To predict the labels, we introduced a simple module that computes a classification of the dependent-governor pairs, obtained as described in \ref{sec:unlabeled}, using their related \textit{context-vectors} as input. If the governor is the root node, the \textit{root vector} is taken instead of the context one. \par

This \textbf{\textit{labeler}} is composed by a simple feedforward network and it is trainined on the gold trees. The training objective is to set the scores of the correct labels above the scores of incorrect ones.\footnote{We use the margin-based objective, aiming to maximize the margin between the highest scoring correct label and the highest scoring incorrect label. We also experimented with the cross-entropy loss activating the output layer with the Softmax, obtaining comparable accuracies but with a slower convergence.} \par

It follows that the \textit{context-encoder} produces a representation that is shared by the \textit{heads-encoder} and the \textit{labeler}, receiving two contributions during the training phase. This sharing of parameters can be seen as an instance of multi-task learning \cite{Caruana:1997:ML:262868.262872}.

As we show in Section \ref{sec:results}, this method is effective: training the \textit{context-encoder} to be good at supporting the prediction of the arc labels significantly improves the convergence of the \textit{heads-encoder},  increasing the global unlabeled attachments score.

\subsubsection{Part-of-Speech Tagger}
\label{sec:tagger}

A typical approach of syntactic parsing assumes that input tokens are morphologically disambiguated using a part-of-speech (POS) tagger before parsing begins. This is bad especially for richly inflected languages (e.g. Italian and German), where there is a considerable interaction between morphology and syntax, such that neither can be fully disambiguated without considering the other. \par

To train a parser for real-world setting, POS tags predicted by an external model are used instead of the gold ones. Modern neural approaches take advantage of pre-trained word embeddings and other token representations (e.g. characters embeddings) to overcome the lack of gold information, achieving results similar to those obtained using gold data. \par

However, most of them focus on dependency parsing without worrying about giving in output POS tags coherent with the predicted labels.\par

Differently, we extend the \textit{labeler} (Section \ref{sec:labeled}) to predict the arc label and the gold coarse-grained part-of-speech jointly, using two feedforward networks that share the same hidden layer. Intuitively, to predict the POS tag of a token the model considers simultaneously the neighboring words, as in most POS taggers, and its syntactic function within the sentence. The label-POS pair is chosen maximizing the sum of the label score and the POS score, evaluating only the pairs seen in the training set.  \par

In this configuration we are experimenting different ways to create the initial tokens encoding (see \ref{sec:expencoding}). \par

\subsection{Virtual Root}
\label{sec:theroot}

Since the dependency parsing relies on a ``verb centricity'' theory \cite{Tesniere:1959}, one of the requirements for a well-formed dependency tree is that there is precisely one root, which is usually the main finite verb of a sentence. \par Our model complies with this by selecting the token connected to the root before any other dependency. \par

The \textit{root vector} is initialized with random values and can be trained only with the labeled parsing (Section \ref{sec:labeled}). When fine-tuned it helps to increase the accuracy of the root attachments. \par

We have tried alternative solutions to avoid having an external root vector. One of these is to force the autoencoder to reconstruct the token itself in case this one points to the root. An extensive benchmark to chose the optimal solution has still to be done.

In addition, we are testing if our model is misled by the garden-path sentences \cite{frazier1982making}, which usually create problems with greedy decoding algorithms.

\section{Latent Syntactic Structure}
\label{sec:lss}

\begin{figure*}[h]
\begin{center}
\includegraphics[scale=1.20]{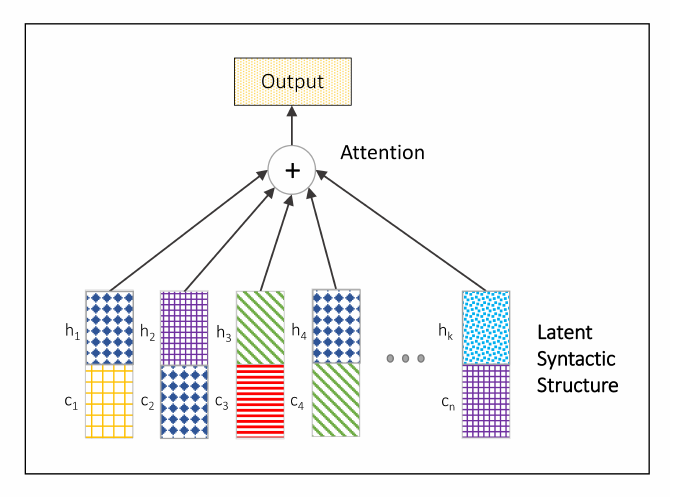}
\end{center}
\vspace*{-1em}
\caption{\small Example of a Latent Syntactic Structure in input to an Attention Mechanism.}
\label{fig:schema_lss}
\end{figure*}

Starting from the results of the neural process described in Section \ref{sec:unlabeled}, it is possible to construct a ``latent syntactic structure'' by concatenating for each \textit{context vector} their related \textit{latent heads}. A way to prepare the input for other semantic tasks is using an attention mechanism \cite{bahdanau2014neural} as a features extractor capable to recognize the relevant information for a given task (Figure \ref{fig:schema_lss}). This makes possible to train the parser together with the task to support, incorporating its objective directly, without requiring the latter to interpret the syntactic output.

However, the focus of this paper is the dependency parsing and not how to use its results, so we leave this topic to future works (see Section \ref{sec:future}).

\section{Experiments and Results}
\label{sec:results}
\textbf{Please note that this section will be updated with new results soon. At the moment it contains only the results that constitute our baseline.}\par

The parser is implemented in Kotlin, using the SimpleDNN\footnote{\url{https://github.com/KotlinNLP/SimpleDNN}} neural networks library. The code is available at the github repository LHRParser.\footnote{\url{https://github.com/GrellaCangialosi/LHRParser}} \par

Rather then top parsing accuracy, in this paper we focus more on the ability of the proposed model to learn a latent representation capable to capture the information needed for the syntactic analysis. \par

A performance evaluation has been carried out on the Penn Treebank (PTB) \cite{penntb} converted to Stanford Dependencies (Marneffe et al., 2006) following the standard train/dev/test splits and without considering punctuation markers. This dataset contains a few non-projective trees. \par

Our baseline is obtained following the labeled parsing approach described in section \ref{sec:labeled}. The input tokens encodings are built concatenating word and POS embeddings (initialized with random values and fine-tuned during the training). The part-of-speech tags are assigned by an automatic tagger.\footnote{The predicted POS-tags are the same used in \newcite{dyer2015transitionbased} and \newcite{kiperwasser2016ef}. We thank Kiperwasser for sharing their data with us.} \par

Like \newcite{kiperwasser2016simple}, during the training we replace the embedding vector of a word with an ``unknown vector'' with a probability that is inversely proportional to the frequency of the word in the tree-bank (tuned with an $\alpha$ coefficient). \par

We optimize the parameters with the Adam \cite{kingma2014adam} update method.\footnote{We use the default parameters ($\alpha$ = 0.001 $\beta$ = 0.9 $\beta$ = 0.999).} \par

The hyper-parameters\footnote{We performed a very minimal tuning of the hyper-parameters.} used for our baseline are reported in Table \ref{tbl:hyper} and the related results in Table \ref{tbl:baseline}.

\begin{table}[ht]
  \begin{center}
    \begin{scalebox}{0.8}{
      \begin{tabular}{ | c | c | }
        \hline 
        Word embedding dimension & 150 \\
        POS tag embedding dimension & 50 \\
        Labeler hidden dimension & 100 \\
        Labeler hidden activation & Tanh \\
        Labeler output activation & Softmax \\
        BiLSTMs activations & Tanh \\
        $\alpha$ (word dropout) & 0.25  \\
        \hline
      \end{tabular}
    }\end{scalebox}
    \caption{Hyper-parameters used for the baseline.}
    \label{tbl:hyper}
  \end{center}
\end{table}

\begin{table}[ht]
    \centering
  \begin{center}
    \begin{scalebox}{0.8}{
      \begin{tabular}{ | c | c | c |  c |}
        System & Method & UAS & LAS \\
        \hline 
        This work (baseline) & LHR & 92.8 & 90.4 \\
        Kiperwasser16 & BiLSTM + transition & 93.2 & 91.2 \\
        Kiperwasser16 & BiLSTM + graph & 93.1 & 91.0 \\
        \hline
      \end{tabular}
    }\end{scalebox}
    \caption{Our results are compared with the models of Kiperwasser and Goldberg (2016), which combine the BiLSTMs with traditional parsing algorithms. Both approaches use neither gold tags nor external resources.}
    \label{tbl:baseline}
  \end{center}
\end{table}

The sections \ref{sec:punct} and \ref{sec:expencoding} detail the other experiments in progress.

\subsection{Punctuation}
\label{sec:punct}

Experimental results with traditional parsing algorithms showed that parsing accuracy drops on sentences which contain a higher ratio of punctuation. The problem is that, in tree-banks, punctuation is not consistently annotated as words, and this makes the learning and the consequent parsing processes difficult. Therefore, the arcs leading to the punctuation tokens just do not count in the standard CoNLL evaluation.\par 

In our model there are no structural constraints to learn the \textit{Latent Heads Representation} so that a fully complete gold dependency tree is not required. Based on this, during the learning process we skip the reconstruction of the head of the punctuation tokens, letting the model creating its own preferred representation. During decoding, the \textit{latent heads} of the punctuation tokens are treated like the others. \par

On the one hand we did not find appreciable improvements applying this method to the PTB, but on the other hand, we did not notice a drop in performance in the evaluation that do not consider the punctuation.\footnote{It could be interesting to find out which tokens the parser chooses as punctuation heads.} This shows a robust behavior of our approach with inaccurate annotations and it supports the hypothesis that the relevant information brought by the punctuation-tokens are implicitly learned thanks to the bidirectional recurrent mechanism \cite{grella2018taking}. \par

We are doing further experiments to test this technique on other tree-banks with a higher ratio of non-projective sentences.

\subsection{Tokens Encoding}
\label{sec:expencoding}

A good initial tokens encoding is crucial to obtain high results in neural parsing. \par We explored different ways to transform the tokens of a sentence into a distributed representation, exploiting the capabilities of our model to predict dependency labels and POS tags jointly (Section \ref{sec:tagger}).

\paragraph{Character-based representation}
\label{sec:charrepr}

\newcite{dozat2017stanford} proof that adding subword information to words embeddings is useful to improve the parsing accuracy, especially for richly inflected languages. \par
We follow their approach, concatenating the word embeddings to a characther-based representation instead of the POS embeddings.

\paragraph{Part-of-Speech Correction}
\label{sec:poscorr}

The input tokens encodings are built concatenating word and POS embeddings. As usual, the part-of-speech tags are assigned by a pre-existent automatic tagger. We use the approach described in Section \ref{sec:tagger} to learn to predict the gold tags together with the dependency label. During the decoding, the output tags of the \textit{labeler} are assigned to each token, eventually modifying the tags predicted by the external POS tagger. The assumption is that our model should learn how to correct the mistakes made by the tagger.

\section{Related Works}
\label{sec:related}
Like us, \newcite{zhang2016dependency} formalized the dependency parsing as the task of finding for each word in a sentence its most probable head. They propose a graph-based parsing model without tree structure constraints (\textit{DeNSe}). It employs a bidirectional LSTM to encode the tokens of a sentence, which are used as features of a feedforward network that estimates the most probable head of each token. In their model, the selection of the head of each token is made independently of the other tokens of the sentence, computing the associative score of all the combinations of pairs of tokens seen as dependent and governor. The most probable pairs are chosen as arcs of the resulting dependency tree, adjusting the ill-formed trees with the Chu-Liu-Edmonds algorithm. \par A key difference between \textit{LHR} and \textit{DeNSe} lies in the training objective: ours is to globally minimize the mean absolute error between the \textit{context vectors} and the \textit{latent heads}, optimizing the BiRNN autoencoder (the \textit{heads-encoder}); by contrast, the objective of \textit{DeNSe} is to minimize the negative log likelihood of the independent predictions of each single arc, respect to the gold arcs in all the training sentences.

\section{Future Research}
\label{sec:future}
In this section we take some space to share a few insight about the direction of our future research:

\paragraph{Multi-objective Training}

We observed meaningful improvements on the unlabeled parsing after adding the labels, using the \textit{context-encoder} as a shared intermediate layer. \par

On this basis, we will try to ``inject'' a linguistic knowledge into the model, involving the same encoder in other known tasks (e.g., semantic role labeling, named entity recognition), and preparing a number of training objective targeted for the improvement of specific difficult dependency relations \cite{ficler2017improving} and to create a ``neuralized'' lexical information (e.g., lemma, grammatical features, valency) contained into computational dictionaries.

\paragraph{Cross-lingual and Unsupervised Dependency Parsing}

In our model there are no structural constraints to learn the \textit{Latent Heads Representation} so that a fully complete gold dependency tree is not required. We plan to experiment with cross-lingual parsing by training the model on large amounts of incomplete and noisy data, obtained by means of annotation projection \cite{hwa2005bootstrapping} or transfer learning.\footnote{Annotation projection is a technique that allows to transfer annotations from one language to another within a parallel corpus.} \par 

In addition, we will experiment new approaches to neural language models \cite{bengio2003neural} based on our bidirectional recurrent autoencoder, with the aim to improve unsupervised parsing techniques \cite{jiang2016unsupervised}.

\paragraph{Semantic Tasks}
We plan to test the effectiveness of our ``latent syntactic structure'' evaluating its contribution to a number of semantic tasks: sentiment analysis (Socher et al., 2013b; Tai et al., 2015), semantic sentence similarity (Marelli et al., 2014), textual inference (Bowman et al., 2015) and neural machine translation (Bahdanau et al., 2015; Jean et al., 2015b).

\section{Conclusion}

The dependency parsing has traditionally been recognized as a structured prediction task. In this paper we have introduced an alternative semi-supervised approach that we believe can radically transforms the way to perform the dependency parsing. \par

To the best of our knowledge, we are the first who use a bidirectional recurrent autoencoder to recognize word-to-word dependencies among arbitrary positions in a sequence of words directly, without involving any additional frameworks.\par

We are investigating what kind of ``knowledge of language'' the new model is capturing, extending the tests to grammaticality judgments and visualizing which information the networks consider more important in a given moment \cite{karpathy2015visualizing}.\footnote{In our experiments we found that the RAN \cite{lee2017recurrent} is a valid alternative to the LSTM when speed and highly interpretable outputs are important.}

\bibliographystyle{acl}

\clearpage

\end{document}